\documentclass[letterpaper, 10 pt, conference]{ieeeconf} 

\IEEEoverridecommandlockouts

\usepackage{graphicx} % for pdf, bitmapped graphics files
\usepackage{amsmath} % assumes amsmath package installed
\usepackage{enumerate}
\usepackage{multirow}
\usepackage{caption}
\usepackage{subcaption}
\usepackage{bbold}

\usepackage[normalem]{ulem}

\usepackage[font={footnotesize}]{caption}
\usepackage[font={footnotesize}]{subcaption}
\usepackage{comment}
\usepackage{booktabs}
\usepackage{gensymb}
\usepackage[linesnumbered, ruled]{algorithm2e}
\usepackage{amsmath}
\usepackage{epstopdf}

\usepackage[utf8]{inputenc}
\usepackage{color}
\usepackage[dvipsnames]{xcolor}
\definecolor{myRed}{HTML}{CF4858}
\definecolor{myGray}{HTML}{4D505B}
\definecolor{myGreen}{HTML}{16A79D}
\definecolor{myPurple}{HTML}{80628B}
\definecolor{myOrange}{HTML}{F4AC42}
\usepackage{mathtools}

\ifodd 1

\else

\fi

\usepackage{cite}

\begin{document}
\title{Depth Completion via Inductive Fusion of Planar LIDAR and Monocular Camera}

\author{Chen Fu$^1$, Chiyu Dong$^1$, Christoph Mertz$^2$ and John M. Dolan$^{1, 2}$ % <-this % stops a space% <-this % stops a space
%\thanks{$*$ Indicates authors contribution equally}
\thanks{$^1$Chen Fu and Chiyu Dong are with the Department of Electrical and Computer Engineering, Carnegie Mellon University, 
          Pittsburgh, PA 15213, USA
        {\tt\small \{cfu1, chiyud\}@andrew.cmu.edu}}%
\thanks{$^2$Christoph Mertz and John M. Dolan are with the Robotics Institute, Carnegie Mellon University, 
         Pittsburgh, PA 15213, USA
        {\tt\small \{mertz, jmd\}@cs.cmu.edu}}%
}

\maketitle
\begin{abstract}
Modern high-definition LIDAR is expensive for commercial autonomous driving vehicles and small indoor robots. %
An affordable solution to this problem is fusion of planar LIDAR with RGB images to provide a similar level of perception capability. %
Even though state-of-the-art methods provide approaches to predict depth information from limited sensor input, they are usually a simple concatenation of sparse LIDAR features and dense RGB features through an end-to-end fusion architecture. %
In this paper, we introduce an inductive late-fusion block which better fuses different sensor modalities inspired by a probability model. %
The proposed demonstration and aggregation network propagates the mixed context and depth features to the prediction network and serves as a prior knowledge of the depth completion. %
This late-fusion block uses the dense context features to guide the depth prediction based on demonstrations by sparse depth features. %
In addition to evaluating the proposed method on benchmark depth completion datasets including NYUDepthV2 and KITTI, we also test the proposed method on a simulated planar LIDAR dataset. %
Our method shows promising results compared to previous approaches on both the benchmark datasets and simulated dataset with various 3D densities.

\end{abstract}

%\IEEEpeerreviewmaketitle

\section{Introduction}\label{sec:intro}
A robust perception system that understands the surrounding scene is crucial for autonomous vehicles. %
Instead of simply detecting the obstacles in 3D world coordinates, this system should provide additional information such as the heading angle, distance, and 3D shape of the obstacles \cite{yeqiang, ChenFu1}. %
To achieve this goal, current autonomous vehicles are usually equipped with high-definition LIDAR (Light detection and Ranging) (e.g. Velodyne LIDAR). %
However, this family of sensors has certain limitations. %
Firstly, high-definition LIDAR is not affordable for commercial autonomous vehicles. %
Even though it provides rich depth information, it lacks detailed texture of the objects, which makes object detection and classification difficult. %
Relying on a single LIDAR is dangerous owing to potential sensor failure. Finally, compared to a RGB image, the point cloud of a LIDAR is sparse, especially for those objects that are far away from the ego-vehicle.%

\begin{figure}[htbp!]
      \centering
      \includegraphics[width = 0.8\linewidth]{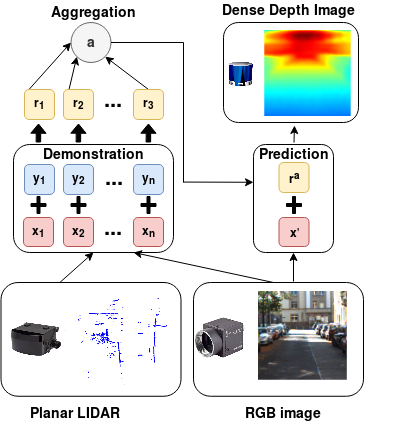}
      \caption{We propose a novel late fusion architecture which better combines dense context features with sparse depth features by inductive fusion block. %
      Our network takes data from affordable RGB camera and planar LIDAR and generates a depth image with similar density to that of high-definition LIDAR. }
      \label{fig:itro_arch}
      \vspace{-0.7 cm}
\end{figure}

Fusing multiple planar LIDAR with an RGB camera is an affordable solution to these limitations. %
This system can provide the autonomous vehicle with sparse 3D information of the surrounding scene and semantic understanding. However, this type of system has a few challenges: %
1) It is difficult to detect and classify objects using the sparse point clouds from planar LIDAR. %
2) Though a single RGB image has rich texture and semantic information, it does not give direct information about the depth of the surrounding environment. 
3) How to fuse the sparse depth information from planar LIDAR with dense RGB semantic features from camera is still an unsolved problem. %
In this paper, we complete a sparse depth map from planar LIDAR by the guidance of semantic and texture information from the RGB image. To achieve this goal, we improve the traditional fusion architecture by introducing an inductive late-fusion technique which achieves a more accurate depth map with more detailed and smoother features, as shown in Figure \ref{fig:itro_arch}. %

We organize this paper as follows: Section \ref{sec:related} briefly reviews the prior work in depth prediction and depth completion. %
Section \ref{sec:method} introduces the proposed network architecture. Section \ref{sec:induct} details the proposed inductive late-fusion network. The experimental setup and dataset are explained in Sections \ref{sec:setup}. Finally, Sections \ref{sec:result_dis} and \ref{sec:conclusion} discuss the experimental results and give conclusions. 

\begin{figure*}[htbp]
      \centering
      \includegraphics[width = 0.95\linewidth]{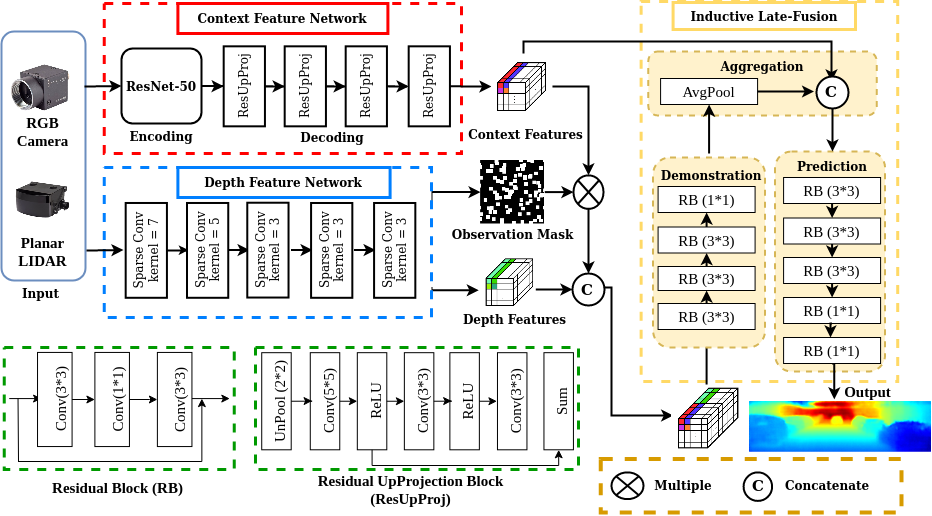}
      \caption{Detailed architecture of proposed inductive late-fusion network. %
      The red and blue blocks indicate the context feature extraction network and depth feature extraction network. %
      The corresponding Residual Block (RB) and Residual UpProjection Block (ReUpProj) are shown in the green blocks. %
      The proposed inductive late-fusion block is shown in the yellow blocks. %
      }
      \label{fig:network}
      \vspace{-0.7cm}
\end{figure*}

\section{Related Work}\label{sec:related}
In robotics perception and especially in autonomous driving, depth estimation is a difficult regression problem. It provides enhanced input for other perception tasks such as vehicle segmentation, tracking and state estimation \cite{ChenFu1}. %
The depth estimation problem can be further divided into different categories according to different sensor inputs. % 
It includes depth in-painting for RGB-D camera and depth completion for planar LIDAR \cite{ma2018self, chenfu2}. % 
\begin{comment}
\textbf{Depth In-painting:}
Due to the sensing limitation of commodity-grade RGB-D cameras, large holes and missing areas exist in the depth image.
This is usually caused by the limited sensing capacity, transparency and brightness of the reflecting surfaces. %
To fill these holes with the Kinect depth image, spatial and temporal information from the nearby pixels is usually considered \cite{doi:10.1117/12.911909,Zhang2013}. %
However, this family of algorithms only considers the local depth information but 3D geometry. %
To tackle this shortcoming, \cite{Matsuo2015} introduces the local tangent plane estimation method into the depth estimation pipeline, which enhances the prediction performance. %
Taking advantage of a deep network which predicts the surface normal, the depth in-painting problem is formulated as a linear optimization problem which takes depth as a regularization term \cite{zhang2018deepdepth}. %
Recent works also apply Generative Adversarial Network (GAN) to in-paint the missing holes within the depth image, which shows better performance \cite{Goodfellow2014}. %
Though RGB-D cameras provide both the RGB image and depth information, they can only sense the depth of the obstacles within a short range, and are therefore not suitable for large-scale outdoor autonomous driving scenarios. % 
\end{comment}

In the dense completion problem, a low-resolution or sparse depth image is completed or super-resolved into a high-quality pixel-wise depth image. 
In this case, more depth points are augmented for the surrounding scenes and objects, given an RGB image and sparse LIDAR point cloud \cite{Held-2013-103059}.  
This depth augmentation algorithm enhances the accuracy of the vehicle tracking task. %
However, this approach only considers the dense completion task as interpolation, which does not predict a pixel-wise dense depth image. %
To further improve the prediction, \cite{Lu2011} adapts a novel data term into the MRF and improves the resolution of the depth image qualitatively and quantitatively. %
As an improvement \cite{Lu2015} considers the relationship between the boundaries of image segmentation and depth image and achieves a better depth prediction for aggressively down-sampled RGB images. %
In order to recover dense depth information from cropped input, \cite{Uhrig2017THREEDV} introduced a novel super-resolved architecture, the Sparsity Invariant Convolution network. %
However, this method is insufficient for the perception system of autonomous vehicles, as multi-modal sensing data need to be fused together. 
To better fuse the input from the RGB image and sparse LIDAR, \cite{Ma2017SparseToDense, ma2018self} proposed a self-supervised training pipeline that takes the sequential information of the RGB image as a geometry constraint of the optimization. % 
However, the performance of this architecture is highly reliant on the accuracy of the transition relation between nearby frames, which can be influenced by moving objects within the scene. %
In contrast, \cite{Qiu_2019_CVPR, Chen_2019_ICCV} fuses the LIDAR and RGB image through deep CNN. With the guidance of the surface normal and intention map, a more accurate depth image is predicted. %
\cite{sddcnn} even proposed an architecture which handles potential sensor failures in real autonomous driving cases. %
However, these depth completion methods\cite{tang2019learning, Qiu_2019_CVPR, Chen_2019_ICCV} take dense LIDAR instead of planar LIDAR as input to predict a pixel-wise depth image. %
Instead of simply concatenating different feature modulations via an end-to-end network architecture, we propose a inductive late-fusion block which fuses the planar LIDAR depth information with context information. % 
This architecture formulates the fusion problem as a conditional distribution. %
With the prior knowledge of the observed LIDAR depth feature and RGB context feature, the underlying distribution of the depth is learnt by an induction framework. %

\section{Method}\label{sec:method}

In this section, we explain in detail how to tackle this sensor fusion problem. %
Compared to state-of-the-art methods which fuse LIDAR depth images with RGB images \cite{tang2019learning, Qiu_2019_CVPR, Chen_2019_ICCV}, fusing sparse (planar) LIDAR and RGB images is quite difficult. %
The challenge is to combine multi-modal features, where the RGB features are dense but the LIDAR features are sparse. %
Instead of fusing these two streams of information by an end-to-end deep network, we propose a novel architecture which fuses the depth and RGB features through induction. %
We combine a depth pathway into our previously proposed depth completion network with inductive fusion units adapted from the conditional neural process technique \cite{pmlr-v80-garnelo18a}. %
This fusion architecture fits the planar LIDAR and RGB image fusion scenario even with an extremely low sparsity. %
Our proposed network has three parts: depth feature extraction network, context feature extraction network and the inductive fusion block. 
The detailed architecture is discussed in the following sections. %
\begin{figure}[htbp!]
      \centering
      \includegraphics[width = 0.9\linewidth]{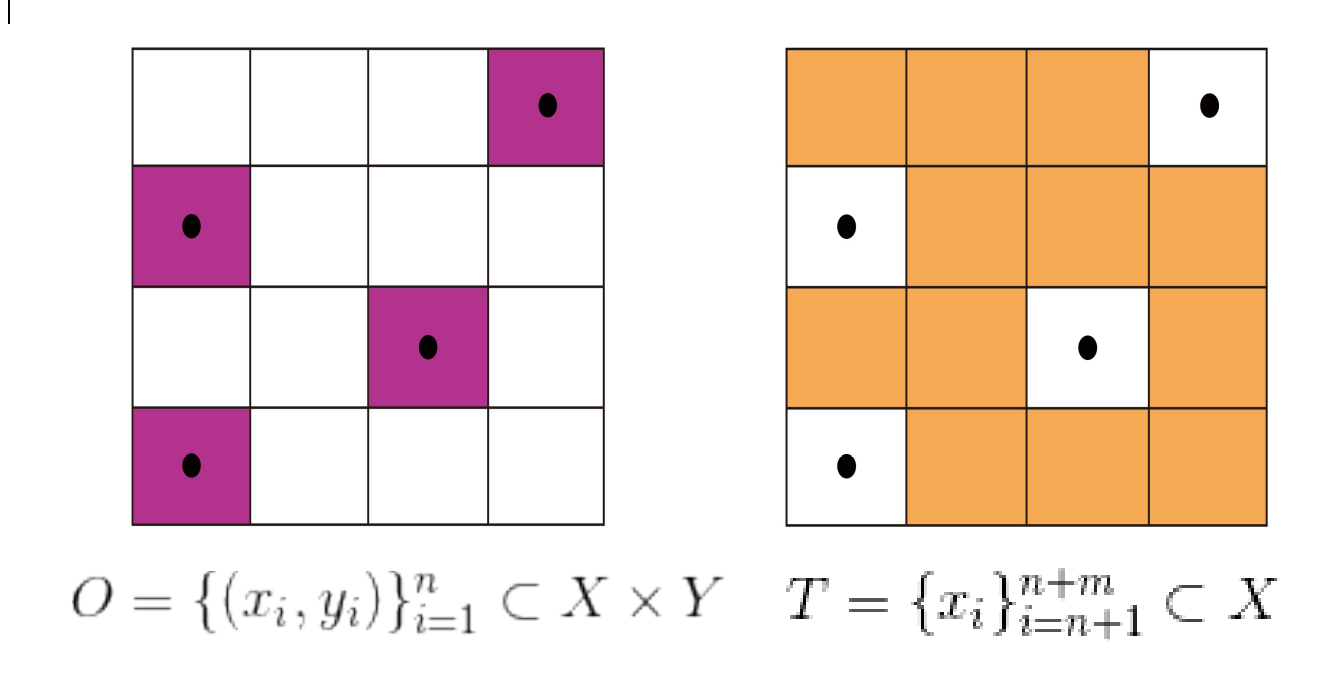}
      \caption{Assuming this is an image plane with a dimension of $4 * 4$, the figure shows an example of the observation set and target set shown in purple and orange respectively. %
      The black dots represent the pixels where we have depth features. }
      \label{fig:cnp_expalin}
      \vspace{-0.7 cm}
\end{figure}
\subsection{Inductive Late-level Fusion}\label{sec:induct}
Due to the sparsity of the input depth image, we cannot directly achieve dense depth features from the network. %
In this case, instead of directly fusing the sparse depth features with dense context features, we first formulate this fusion problem by a probability model. %
We currently have two sets of input, the observation set (N = $n$) and target set (N = $m$), where $n+m$ equals the total number of pixels in the image. %
In the observation set, each pixel location has a context feature vector and a depth feature vector associated with it. %
However, in the target set, we only have the context features.
We define the observation set as $O = \{(x_i, y_i)\}_{i = 1}^{n} $, where $x_i$ represents the context feature vector and $y_i$ represents the depth feature generated by the feature extraction network. %
The target set is $T= \{x_i\}_{i =n+1}^{n + m}$. %
An example observation set and target set is shown in Fig. \ref{fig:cnp_expalin}. %
We now define a mapping function $f: X \xrightarrow{} Y$ and let $P_\theta$ be a probability distribution over the function $f$. %
For $f \sim P_{\theta}$ we set $\hat{y} = f(x_i)$, where $\hat{y}$ is the ground truth depth for each pixel. %
So $P_\theta$ defines a joint probability distribution over $\{x_i\}_{i =n+1}^{n + m}$ and the depth completion problem is formulated as a Gaussian Process (GP) as follows: %
\begin{equation}\label{equ:dp:gp}
    P_\theta(f(T)|O, T)
\end{equation}
To solve this GP problem, an appropriate prior knowledge is required. %
However, it is difficult to model the complex correlation structure between depth value and RGB image using a simple kernel function. %
Moreover, the GP is computationally expensive, as the scale depends on the input and output dimension, which is large in the depth completion problem. %
In this case, we created a novel adaptation of the Conditional Neural Process (CNP) \cite{pmlr-v80-garnelo18a} architecture that addresses the above limitations. %

In equation \ref{equ:dp:gp}, $\theta$ represents all parameters of the inductive fusion block, which defines the distribution $P$. %
In the depth completion task, the pixels in each image patch do not have the problem of ordering. %
In other words, the depth value of a particular pixel is highly dependent on the surrounding pixels but not the order of the pixels. %
As a result, the prediction is permutation-invariant and has the following property, which fits the assumption of CNPs:
\begin{equation}\label{equ:permutation}
    P_\theta(f(T) | O, T) = P_\theta(f(T') | O, T') = P_\theta(f(T) | O', T)
\end{equation}
where $O'$ and $T'$ are permutations of $O$ and $T$. %
So, we can apply the following equations from CNPs:
\begin{equation}\label{equ:cnp_demo}
    r_i = h_\theta(x_i, y_i), ~~~ \forall(x_i, y_i) \in O
\end{equation}
\begin{equation}\label{equ:cnp_aggreate}
    r_j = \frac{1}{T} \sum_{i=1}^{T} r_i
\end{equation}
\begin{equation}\label{equ:cnp_pred}
    \phi_i = g_\theta(x_i, r), ~~~ \forall (x_i) \in T
\end{equation}
So, the inductive fusion block contains three subnets, the demonstration network, aggregation network and prediction network. %
For the demonstration network, as shown in \ref{equ:cnp_demo}, it fuses the depth feature and context features and encodes the information from two feature extraction networks for each pixel location and achieves a higher-level description of the prior knowledge of the regression task. %
Instead of directly passing the higher-level descriptors to the decoder network, which is also the prediction network, we aggregate them into an intermediate feature representation $r_j$ for each image patch. %
By virtue of the aggregation network, the features are invariant to translation and rotation, which results in a robust network. %
In the prediction network, target depth values are estimated based on aggregated feature and context feature.%

\subsection{Depth Feature Extraction Network}\label{sec:lidar_feature}
To better extract the features from sparse depth image input, we apply the Sparsity Invariant CNN (SCNN) \cite{Uhrig2017THREEDV} to generate the feature representation of the depth. %
Different from the vanilla convolution layer, SCNN introduces a sparse convolution layer which propagates the features as well as an observation mask along the forward path. %
The sparse depth value is diluted to the nearby pixels and results in blurred predictions. %
During this process, the observation mask is also expanded by the max-pooling layer. %
To generate a depth feature for each pixel, we delete the last layer from the network in the proposed model. %
The detailed network architecture is shown in the blue blocks in Fig. \ref{fig:network}. %
\subsection{Context Feature Extraction Network}\label{sec:contect_feature}
For the context feature extraction network, we follow our previously proposed depth prediction architecture described in \cite{chenfu2}. %
In this architecture, we concatenate the popular Resnet model-proposed Residual Up-Projection blocks shown in the green blocks of Fig. \ref{fig:network}. %
This model takes advantage of residual learning techniques and captures more detailed features from the input RGB image features \cite{He2015}. %
In the depth completion task, we find that the detailed features within the depth images show similarity to the low-level features in the RGB images. %
For example, the RGB image and depth image have similar edges and corners. %
So, low-level features from the shallow network contain useful features for the depth prediction and reconstruction in higher layers. %
To further improve the prediction results, we add three skip connections to pass low-level features to the residual Up-Projection blocks. %
Instead of using the whole network as a context feature extraction network, we get rid of the top layers and the network outputs a context feature tensor. %
\begin{table*}[h!]
%% increase table row spacing, adjust to taste

\renewcommand{\arraystretch}{1}
% if using array.sty, it might be a good idea to tweak the value of
%\extrarowheight %as needed to properly center the text within the cells
\caption{Performance comparison of proposed network with previous single RGB-based and fusion methods on NYUdepthV2 dataset with 200 depth points}%We also compare the performance of single RGB-based network with depth and camera fusion network}
\label{tab:msre_nyu}
\centering
\scalebox{1}{
% Some packages, such as MDW tools, offer better commands for making tables
% than the plain LaTeX2e tabular which is used here.
\begin{tabular}[!htbp]{ccccccccc}
\toprule

Input & \#Sample & Methods & RMSE (m) & REL & $\delta_1~(\%$) & $\delta_2~(\%)$ & $\delta_3~(\%)$ \\
\cmidrule(r){1-1}\cmidrule(lr){2-2}\cmidrule(lr){3-3} \cmidrule(lr){4-4}\cmidrule(lr){5-5}\cmidrule(lr){6-6}\cmidrule(lr){7-7}\cmidrule(lr){8-8}\\
RGB &0  & Roy et al.\cite{Roy} & 0.744 & 0.187 & - & - & -\\
 & 0  & Eigen et al.\cite{Eigen14predictingdepth} & 0.641 & 0.158 & 76.9 & 95.0 & 98.8\\%Eigen14predictingdepth
&0   & Laina et al.\cite{laina2016deeper} & 0.573 & 0.127 & 81.1 & 95.3 & 98.8\\%\cite{laina2016deeper}
\midrule 
RGB+D&225    & Liao et al.\cite{liao2017ICRA} & 0.442 & 0.104 & 87.8 & 96.4 & 98.9\\%\cite{liao2017ICRA}
&200    & Ma et al.\cite{Ma2017SparseToDense} & 0.230 & 0.044 & 97.1 & 99.4 & 99.8\\%\cite{Ma2017SparseToDense}]
&200 & Lee et al. \cite{Lee-2019-111938} & 0.225 & 0.046 & 97.2 & - & -\\
&200    & Fu et al.\cite{chenfu2} &  0.203 & 0.040 & 97.6 & 99.5 & $\mathbf{99.9}$\\ 
&200    & Abdelrahman et al. \cite{eldesokey2018propagating} &0.192   & 0.030 &97.9 &99.5 &99.8 \\ 

&200 &  Proposed & $\mathbf{0.169}$ & $\mathbf{0.028}$  & $\mathbf{98.4}$ & $\mathbf{99.9}$ & $\mathbf{99.9}$\\
%&200 & LRP & 0.159 & $0.030$ & $\mathbf{98.6}$ & $\mathbf{99.9}$ & $\mathbf{99.9}$ \\

%\midrule
%&200 & Normal Iter 1.  & 0.209 & 0.046 & 97.5 & 99.6 & $\mathbf{99.9}$\\
%&200 & Normal Iter 1 + AD.&0.186 & 0.038 &97.9 &99.5 &99.8 \\ 
%&200 & Normal Iter 2.  & 0.180 & 0.036 & 98.1 & 99.6 & $\mathbf{99.9}$\\
%&200 & Normal Iter 2 + AD.&0.175 & 0.034 &98.3 &99.7 & $\mathbf{99.9}$ \\
%&200 & Normal Iter 3 &0.174 & 0.034 &98.3 &99.7 & $\mathbf{99.9}$ \\
%\midrule
%&200 & hourglass Iter 1 & 0.185 & 0.037 & 98.0 & 99.6 & 99.9 \\
%&200 & hourglass Iter 1 + AD & 0.167 & 0.032 & 98.5 & 99.7 & 99.9 \\
%&200 & hourglass Iter 1 + DFT + AD & 0.165 & 0.030 & 98.5 & 99.7 & 99.9 \\
%&200 & hourglass Iter 2 & $0.161$ & $0.030$ & $\mathbf{98.6}$ & $99.8$ & $\mathbf{99.9}$ \\
%&200 & hourglass Iter 2 + AD & 0.159 & $0.030$ & $\mathbf{98.6}$ & $\mathbf{99.9}$ & $\mathbf{99.9}$ \\
%&200 & hourglass Iter 2 + DFT + AD & $\mathbf{0.157}$ & $\mathbf{0.029}$ & $\mathbf{98.6}$ & $\mathbf{99.9}$ & $\mathbf{99.9}$ \\
%&200 & hourglass Iter ３& $0.161$ & $0.034$ & $98.5$ & $99.8$ & $\mathbf{99.9}$ \\
%&200 & hourglass Iter ４ & $0.163$ & $0.038$ & $98.5$ & $99.8$ & $\mathbf{99.9}$ \\
\bottomrule
\end{tabular}}
%\vspace{-0.5cm}
\end{table*}

\section{Results and Discussion}\label{sec:result}
\begin{table*}[h!]
%% increase table row spacing, adjust to taste
\renewcommand{\arraystretch}{1}
% if using array.sty, it might be a good idea to tweak the value of
%\extrarowheight %as needed to properly center the text within the cells
\caption{Performance comparison of proposed network with previous single RGB-based and fusion methods on KITTI Odometry dataset}
\label{tab:msre_kitti}
\centering
\scalebox{1}{
% Some packages, such as MDW tools, offer better commands for making tables
% than the plain LaTeX2e tabular which is used here.
\begin{tabular}[h!]{cccccccc}
\toprule
Input & \#Sample & Methods & RMSE (m) & REL & $\delta_1~(\%)$ & $\delta_2~(\%)$ & $\delta_3~(\%)$ \\
\cmidrule(r){1-1}\cmidrule(lr){2-2}\cmidrule(lr){3-3} \cmidrule(lr){4-4}\cmidrule(lr){5-5}\cmidrule(lr){6-6}\cmidrule(lr){7-7}\cmidrule(lr){8-8}\\
RGB &0 & Laina et al.\cite{Roy} & 8.73 & 0.280 & 60.1 & 82.0 & 92.6\\
&0  & Mancini et al.\cite{mancini_depth} & 7.51 & - & 31.8 & 61.7 & 81.3\\%mancini_depth
&0   & Eigen et al.\cite{Eigen14predictingdepth} & 6.16 & 0.190 & 69.2 & 89.9 & 96.7\\
\midrule %Eigen14predictingdepth
RGB+D &200    & Liao et al.\cite{liao2017ICRA} & 4.50 & 0.113 & 87.4 & 96.0 & 98.4\\%
&200    & Ma et al.\cite{Ma2017SparseToDense} &3.85 &0.083 &91.9 &97.0 & $\mathbf{98.9}$ \\ %Ma2017SparseToDense
&200    & Fu et al. \cite{chenfu2} & 3.67 & 0.072  & 92.3 & 97.3 & $\mathbf{98.9}$\\
&200    & Proposed & $\mathbf{3.11} $ & $\mathbf{0.058}$  & $\mathbf{93.9}$ & $\mathbf{97.6}$ & $\mathbf{98.9}$\\
%&200    & LRP & $\mathbf{2.46} $ & $0.047$  & $\mathbf{97.3}$ & $\mathbf{99.2}$ & $\mathbf{99.7}$\\
\midrule
RGB+D&500    & Ma et al \cite{Ma2017SparseToDense}& $3.38$ & $0.073$  & $93.5$ & $97.6$ & $98.9$\\
&500    & Cheng et al. \cite{cheng2018depth} & $2.98$ & $\mathbf{0.044}$  & $\mathbf{95.7}$ & $98.0$ & $99.1$\\
&500    & Proposed & $\mathbf{2.84}$ & $0.045$  & $95.3$ & $\mathbf{98.1}$ & $\mathbf{99.2}$\\
%&200    & LRP & $\mathbf{2.40} $ & $0.043$  & $\mathbf{97.8}$ & $\mathbf{99.4}$ & $\mathbf{99.8}$\\
\bottomrule
\end{tabular}}
\vspace{-0.2 cm}
\end{table*}
In this section, we concentrate on explaining our experimental setup and results. %
To verify the performance of the proposed architecture and compare to the state of the art, we evaluate our model on two public datasets: the indoor scenes NYUdepthV2 dataset and the KITTI depth completion dataset, which focuses on real on-road scenes. %

\subsection{Network Implementation and Training Strategy}\label{sec:training}
For the demonstration network, we concatenate 4 Residual Blocks with kernel size of 3 * 3. In the prediction network, we concatenate 5 Residual Blocks and finally predict the depth map. The detailed architecture is shown in Fig. \ref{fig:network}.  
To train the network, we start from the ImageNet pretrained Resnet model and fine-tune the context feature network by the NYUDepthV2 and KITTI dataset. %
For the depth feature extractor, we pretrain the network by the NYUDepthV2 and KITTI dataset as well. %
Finally, we combine the depth feature network with the context feature network through the inductive fusion unit. %
In order to merely compare the network architecture with previous works, we use the L-1 norm as our loss function throughout the training procedure, both the pre-training and whole network architecture. %
The whole network architecture is trained by the SGD optimizer \cite{adam}, with the default parameter setting from the Pytorch optimization package \cite{paszke2017automatic}. %
\begin{comment}
Due to limited computation resources, we use a smaller batch size of 8 and train the network for 30 epochs. %
The learning rate starts from $0.01$ with a decreasing rate of $0.1$ and step size of $5$. %
In order to improve the generalization of the network, we applied an online data augmentation strategy. %
The original input image and depth image are randomly transformed as follows:
\newline
\textbf{Scale:} RGB and depth images are randomly scaled by a factor $S \in [1.0, 1.5]$ according to a uniform distribution. %
The depth image is divided by $s$. %
\newline
\textbf{Flips:}The RGB and depth images are horizontally flipped with a probability of $50\%$. %
\newline
\textbf{Rotation:} We rotate the RGB and depth images by a random angle $\theta \in [-5\degree, 5\degree]$. %
\newline
\textbf{Color transformation:} We normalize the RGB image by subtracting channel mean and dividing by channel standard deviation. % 
\end{comment}

\subsection{Experiment Setup and Evaluation Metric}\label{sec:setup}

\textbf{Dataset}: 
We conduct the experiments on several benchmark datasets introduced by state-of-the-art methods, including NYUDepthv2, KITTI Odometry and the reduced-resolution KITTI dataset \cite{Ma2017SparseToDense, chenfu2}. We apply the same experiment setup as in our previous paper \cite{chenfu2}, with data augmentation including scale, flip, rotation and color translation \cite{Ma2017SparseToDense}.
We apply a uniform sampling strategy to generate different input densities for evaluation. % 

\textbf{Evaluation Metrics}:
In order to compare the proposed method with benchmark results of the state-of-the-art methods, we apply the following standard evaluation metrics. %
We not only directly measure the average error over all pixels by root mean square error (RMSE), but also compare the Mean Absolute Relative Error (REL), which prevents the scaling problem, as follows: %
\begin{equation} \label{equ:REL}
  \mathbf{e}_{rel} = \frac{1}{N}\sum_i\frac{|\hat{y_i} - y_i|}{y_i}
\end{equation}
In order to count the percentage of pixels within a certain threshold, we also consider the $\delta_j$ metric, defined as follows:
\begin{equation} \label{equ:delta}
  \mathbf{\delta}_{j} = \frac{\mathbf{count}(\{y_i: max(\frac{\hat{y_i}}{y_i}, \frac{y_i}{\hat{y_i}}) \leq 1.25^j\})}{\mathbf{count}(y_i)}
\end{equation}
where $\hat{y_i}$ and $y_i$ are the prediction and ground truth depth for each pixel. %
As a result, the method performs better if it achieves a low RMSE and REL or a higher $\delta_j$. %
\begin{figure}[h!]
      \centering
      \includegraphics[width = 1\linewidth]{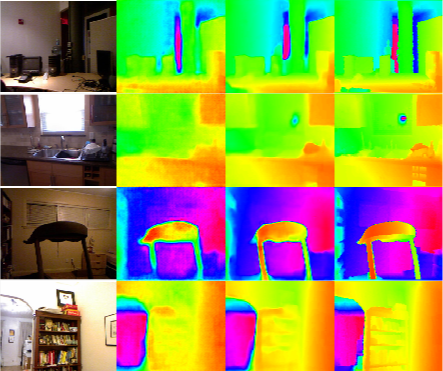}
      \caption{Visualization result on NYUDepthV2 dataset. %
      The images from left to right are: RGB image, prediction results from \cite{chenfu2}, prediction results by proposed method and ground truth. %
      We are using 200 depth points in this experiment setting. %
      }
      \label{fig:result_nyudepth}
      \vspace{-0.7 cm}
\end{figure}
\subsection{Ablation Study}\label{sec:ablation_study}
\begin{figure}[htbp!]
      \centering
      \includegraphics[width = 1\linewidth]{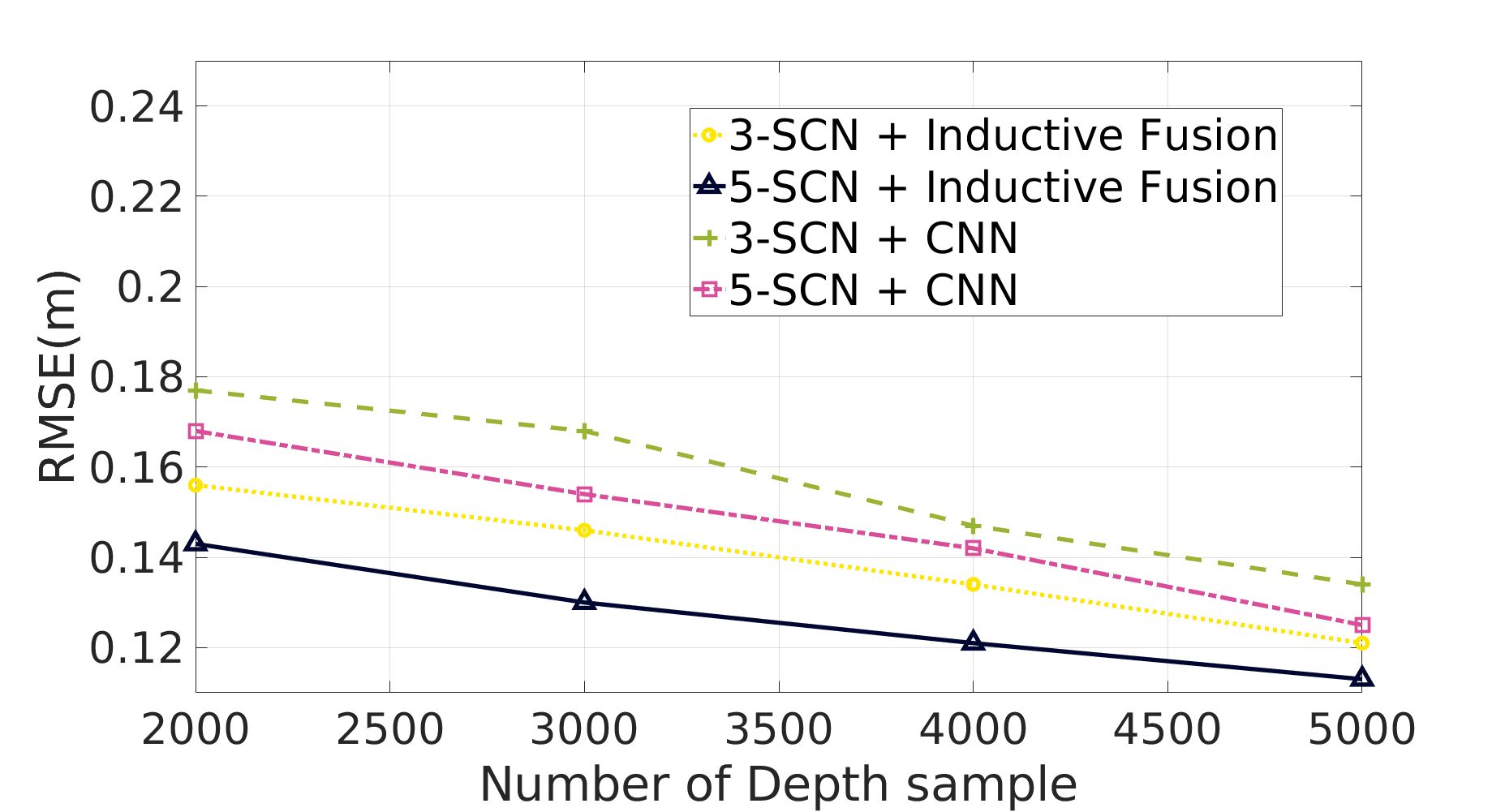}
      \caption{Ablation study of different network architectures and performance with various input 3D densities.%
       In this figure, the proposed late-fusion block is compared with vanilla convolution networks. %
      We also study how different size of depth feature network influences the performance of depth estimation.%
      }
 \label{fig:ablation_study}
  \vspace{-0.5cm}
\end{figure}

In order to study the function of the proposed late-fusion block and prove its advantage, we also provide an ablation study of different fusion architectures and their performance over different input densities. %
We compare our late-fusion block with a simple late-fusion network that uses 9 residual blocks as the fusion network.
We also change the size of the depth feature network by varying the number of sparse convolution layers (SCN). %
In this experiment, we compare the depth feature network with 3 SCN and 5 SCN. %
The result is shown in Fig. \ref{fig:ablation_study}. 

In comparison, the context feature network only achieves a RMSE of 0.46m with a single RGB image as input. In this case, the fusion method has much better performance than the single RGB network. %
Even with an early-fusion input as in \cite{chenfu2}, the context feature network only achieves a RMSE of 0.20m on the NYUDepthV2 dataset. %
As a result, the proposed late-fusion architecture benefits the performance of the depth completion task. %
As we can see in Fig. \ref{fig:ablation_study}, as we increase the number of input depth samples, we achieve better performance and the proposed inductive late-fusion block outperforms the traditional late-fusion network concatenated by convolution layers. %
As the prediction network is estimating the depth value based on the demonstration and observed context features, it gives smoother prediction at image pixels which have similar RGB patterns and geometry constraints. %
In other words, pixels with similar context should have a similar depth value. %
Moreover, pixels that are physically nearby but have distinct context features should have different depth values. % 
As a result, we achieve a smoother prediction on flat surfaces and sharper edges on contours of objects.   %
Instead, traditional network architectures are merely learning the distribution of depth based on RGB image and depth samples, which results in a blurred prediction on both surfaces and edges. %
\begin{comment}
\begin{figure}[htbp!]
  \begin{subfigure}[b]{0.24\textwidth}
    \includegraphics[width=\textwidth]{figures/cnn.png}
    \caption{Vanilla CNN block}
    \label{fig:1}
  \end{subfigure}
  %
  \begin{subfigure}[b]{0.24\textwidth}
    \includegraphics[width=\textwidth]{figures/proposed.png}
    \caption{Inductive Late-fusion block}
    \label{fig:2}
  \end{subfigure} 
  \caption{Ablation study: qualitative comparison of proposed late-fusion block with traditional fusion network with simple convolution layers. %
  In this comparison, we have 200 depth samples.  %
  \label{fig:ablation_study_vis}
      }
      %\vspace{-0.5cm}
\end{figure}
\end{comment}

\subsection{Results and Discussion on Benchmark Dataset}\label{sec:result_dis}
In this section, we mainly discuss the experimental results on the benchmark dataset, NYUDepthV2 and KITTI dataset. %
We take the reported accuracy of state-of-the-art depth completion and depth prediction methods from their original papers. %

\textbf{NYUDepthV2 Dataset}:
In general, taking multi-modal sensor data as input improves the estimation result compared to using a single sensor. %
In methods \cite{Ma2017SparseToDense, liao2017ICRA}, these architectures apply the end-to-end network, which concatenates multiple residual blocks and deep decoder networks. %
However, this family of methods loses detailed features and context information through the encoding and decoding architecture. %
On the NYUDepthV2 dataset, we achieve a $12.5\%$ improvement with 200 dense inputs in terms of RMSE. %
The detailed comparison is shown in TABLE \ref{tab:msre_nyu}. %
%Similarly, \cite{Ma2017SparseToDense} applied the up-projection blocks from \cite{laina2016deeper} and improves the RMSE by $50\%$. %
By applying our proposed architecture, we achieve an $11.9\%$ better performance compared with the baseline method \cite{eldesokey2018propagating} on the NYUDepthV2 dataset. %
In Fig. \ref{fig:result_nyudepth}, we show the visualization result on the NYUDepthV2 dataset. %
Comparing with ground truth, the prediction results of the proposed method have much clearer context details. %
In all, combining REL and RMSE in TABLE \ref{tab:msre_nyu} and Fig. \ref{fig:result_nyudepth}, we achieve a better estimation on the boundaries as well as flat surfaces in the scenes. % 

\begin{figure*}[htbp!]
      \centering
\begin{minipage}{0.245\textwidth}
\begin{subfigure}{\linewidth}
    \centering
    \includegraphics[width=\linewidth]{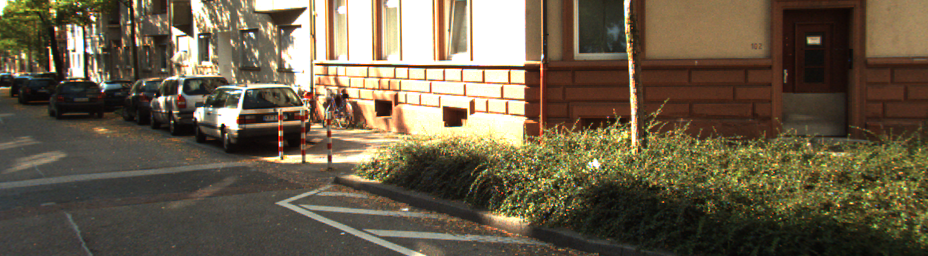}
\end{subfigure}\\[0.2ex]
\begin{subfigure}{\linewidth}
    \centering
    \includegraphics[width=\linewidth]{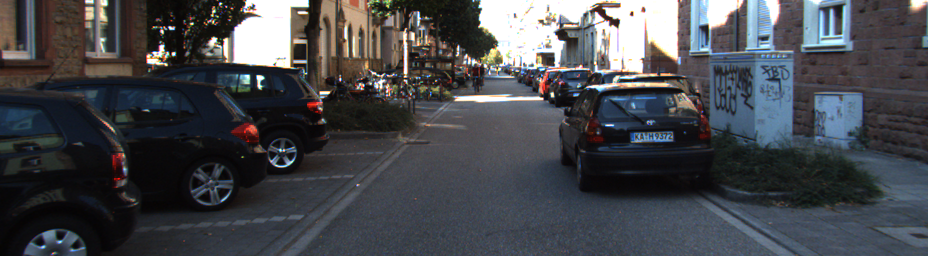}
\end{subfigure}\\[0.2ex]
\begin{subfigure}{\linewidth}
    \centering
    \includegraphics[width=\linewidth]{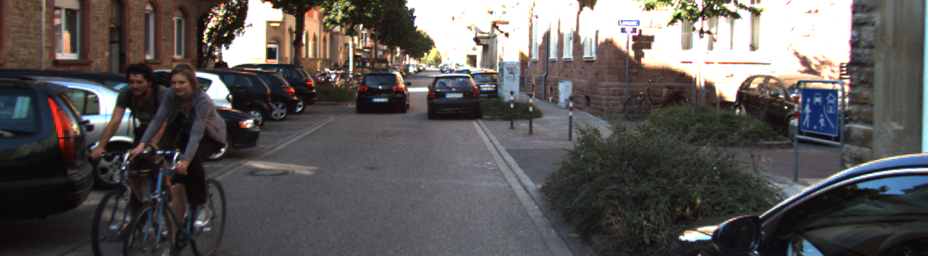}
\end{subfigure}\\[0.2ex]
\begin{subfigure}{\linewidth}
    \centering
    \includegraphics[width=\linewidth]{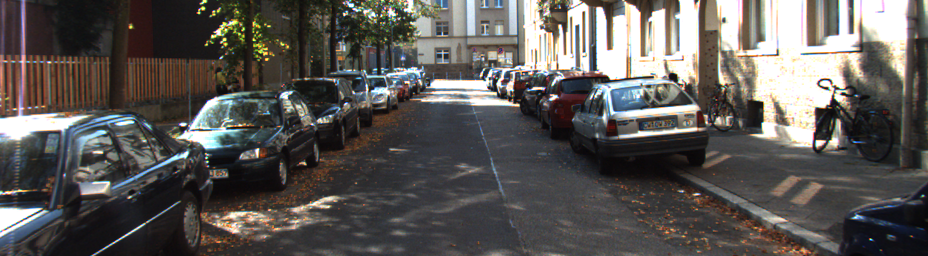}
    \caption{RGB image}
\end{subfigure}\\[0.2ex]
\end{minipage}
\begin{minipage}{0.245\textwidth}
\begin{subfigure}{\linewidth}
    \centering
    \includegraphics[width=\linewidth]{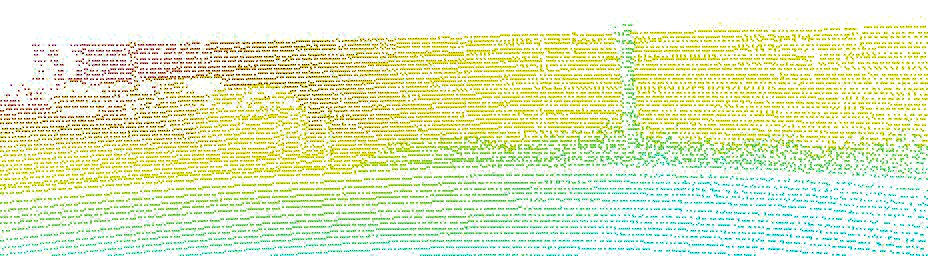}
\end{subfigure}\\[0.2ex]
\begin{subfigure}{\linewidth}
    \centering
    \includegraphics[width=\linewidth]{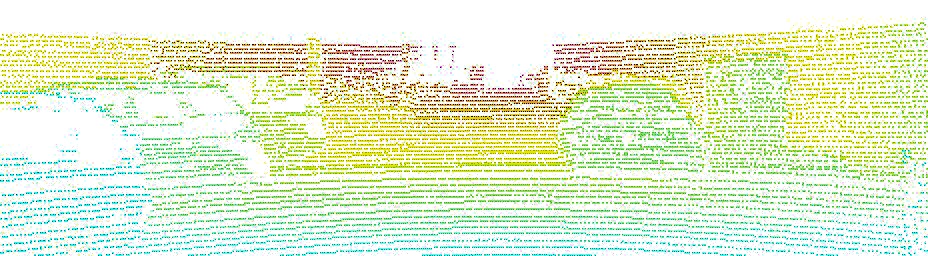}
\end{subfigure}\\[0.2ex]
\begin{subfigure}{\linewidth}
    \centering
    \includegraphics[width=\linewidth]{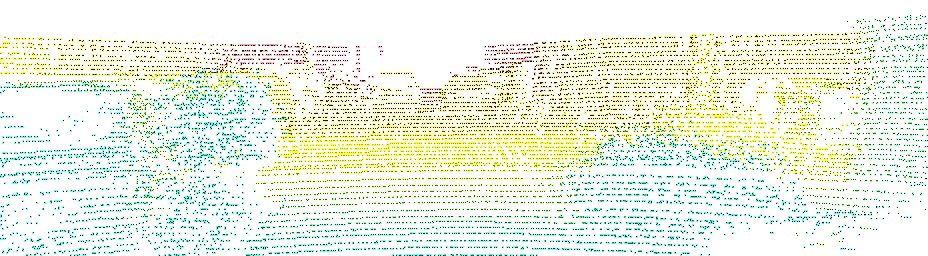}
\end{subfigure}\\[0.2ex]
\begin{subfigure}{\linewidth}
    \centering
    \includegraphics[width=\linewidth]{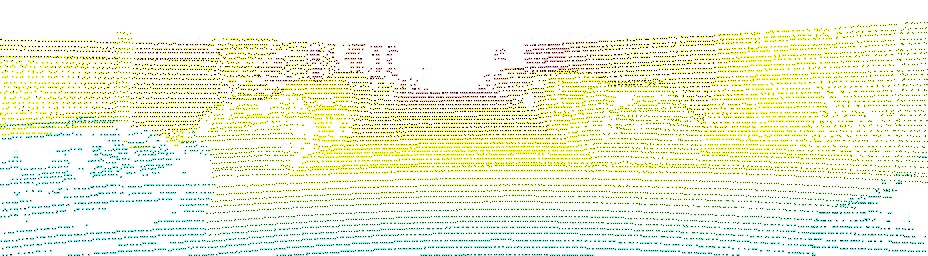}
    \caption{Ground Truth}
\end{subfigure}\\[0.2ex]
\end{minipage}
\begin{minipage}{0.245\textwidth}
\begin{subfigure}{\linewidth}
    \centering
    \includegraphics[width=\linewidth]{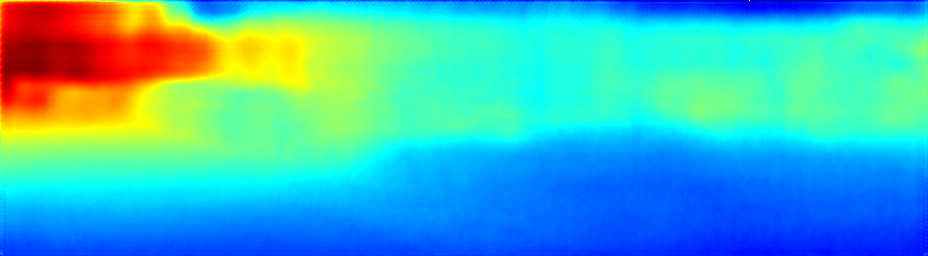}
\end{subfigure}\\[0.2ex]
\begin{subfigure}{\linewidth}
    \centering
    \includegraphics[width=\linewidth]{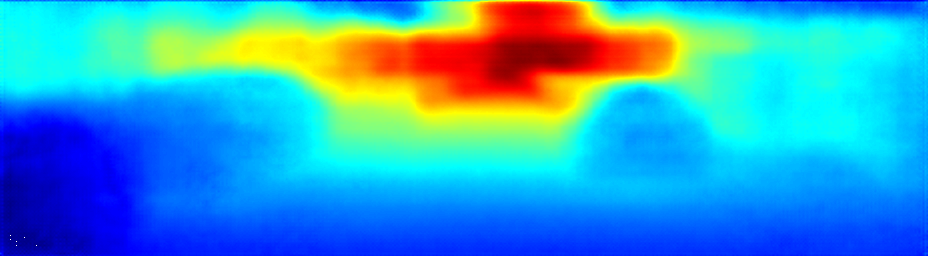}
\end{subfigure}\\[0.2ex]
\begin{subfigure}{\linewidth}
    \centering
    \includegraphics[width=\linewidth]{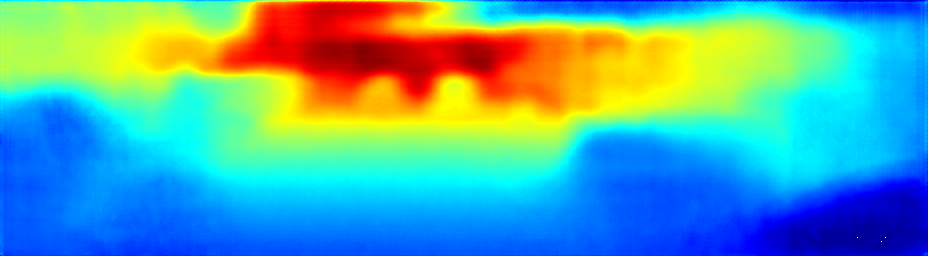}
\end{subfigure}\\[0.2ex]
\begin{subfigure}{\linewidth}
    \centering
    \includegraphics[width=\linewidth]{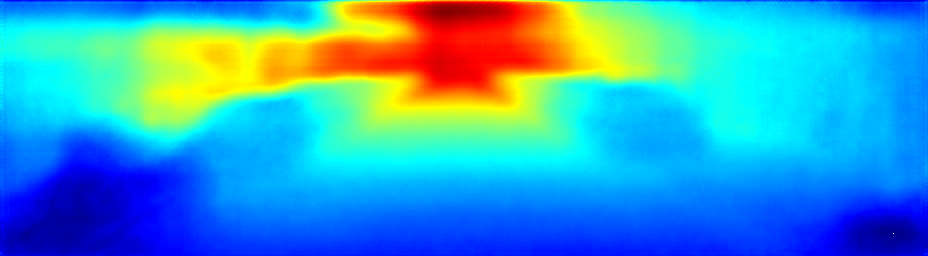}
    \caption{\cite{chenfu2}}
\end{subfigure}\\[0.2ex]
\end{minipage}
\begin{minipage}{0.245\textwidth}
\begin{subfigure}{\linewidth}
    \centering
    \includegraphics[width=\linewidth]{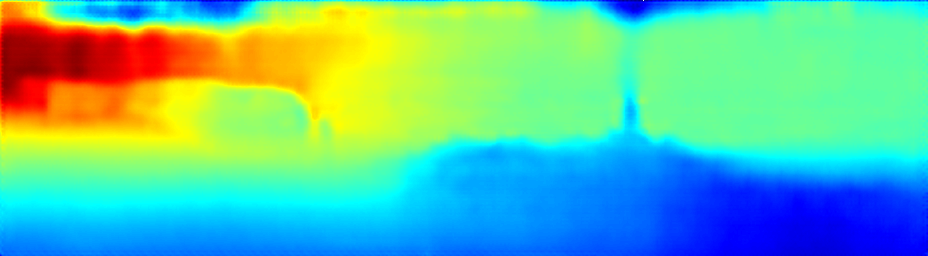}
\end{subfigure}\\[0.2ex]
\begin{subfigure}{\linewidth}
    \centering
    \includegraphics[width=\linewidth]{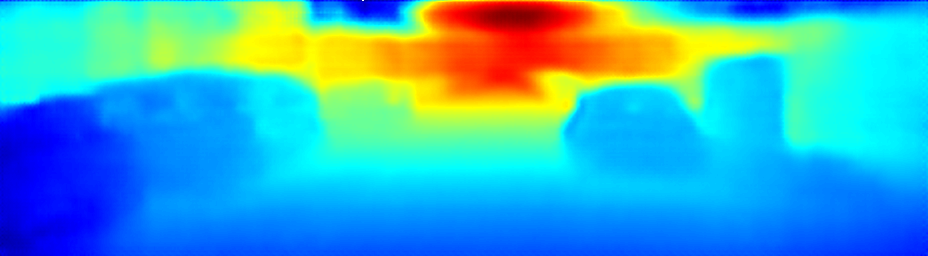}
\end{subfigure}\\[0.2ex]
\begin{subfigure}{\linewidth}
    \centering
    \includegraphics[width=\linewidth]{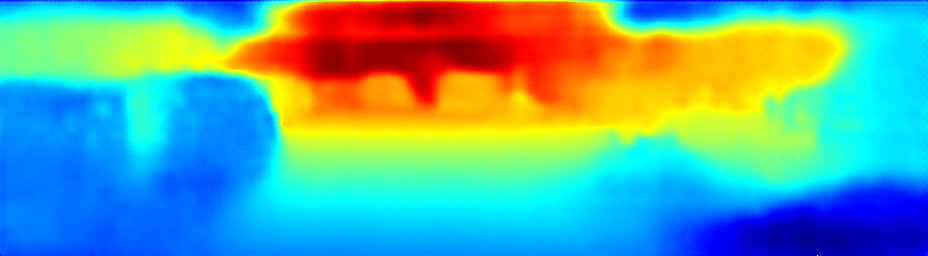}
\end{subfigure}\\[0.2ex]
\begin{subfigure}{\linewidth}
    \centering
    \includegraphics[width=\linewidth]{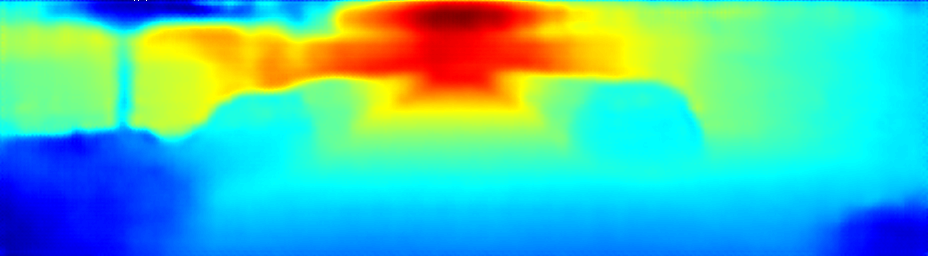}
    \caption{Proposed}
\end{subfigure}\\[0.2ex]
\end{minipage}
      \caption{Visualization result on KITTI dataset on official validation sets. %
      The images from left to right are: RGB image, ground truth dense depth (brightness, contrast changed for visual enhancement), predicted dense depth image from previous work \cite{chenfu2} and prediction dense depth image by proposed method. %
      We are using 200 depth points in this experiment setting. %
      }
      \label{fig:result_kitti}
      \vspace{-0.3 cm}
\end{figure*}

\textbf{KITTI Odometry Dataset and Reduced-Resolution KITTI Dataset}:
In TABLE \ref{tab:msre_kitti}, we notice that fusion of sparse LIDAR input with dense RGB image has better depth completion results. %
In comparing with our previous best depth completion result using 200 depth samples as input, we achieve a $14.7\%$ improvement in terms of RMSE. %
The visualization result is shown in Fig. \ref{fig:result_kitti}. %
We also conducted a performance comparison on the reduced-resolution KITTI dataset and compared with a few state-of-the-art single RGB images with LIDAR fusion algorithms. %
We take 200 depth points to simulate the density of planar LIDAR in the band. %
As we see from TABLE \ref{tab:kitti_band}, the proposed method performs better than previous methods by around $20\%$ and $25\%$ regarding RMSE and REL, respectively. %

\begin{table}[h!]
%% increase table row spacing, adjust to taste
\renewcommand{\arraystretch}{1}
% if using array.sty, it might be a good idea to tweak the value of
%\extrarowheight %as needed to properly center the text within the cells
\caption{Performance comparison of proposed network on reduced-resolution KITTI odometry dataset. %
We take a band of the LIDAR point cloud and conduct depth completion within the band. %
We down-sample the dense point cloud by randomly selecting 200 LIDAR points as input. %
}
\label{tab:kitti_band}
\centering
\scalebox{0.9}{
% Some packages, such as MDW tools, offer better commands for making tables
% than the plain LaTeX2e tabular which is used here.
\begin{tabular}[tbp!]{cccccccc}
\toprule
Methods  & RMSE (m) & REL & $\delta_1~(\%)$ & $\delta_2~(\%)$ & $\delta_3~(\%)$ \\
\cmidrule(r){1-1}\cmidrule(lr){2-2}\cmidrule(lr){3-3} \cmidrule(lr){4-4}\cmidrule(lr){5-5}\cmidrule(lr){6-6}\cmidrule(lr){7-7}\\

      Ma et al.\cite{Ma2017SparseToDense} & 0.42 & 0.040 & 98.3 & 99.6 & 99.8\\
      Fu et al. \cite{chenfu2}& 0.35 & 0.032 & 98.5 & 99.6 & 99.8 \\
       Proposed &  $\mathbf{0.28}$ & $\mathbf{0.023}$ & $\mathbf{98.9}$ & $\mathbf{99.7}$ & $\mathbf{100.0}$\\
\bottomrule
\end{tabular}}
\vspace{-0.5 cm}
\end{table}

\section{Conclusion}\label{sec:conclusion}
In this paper, we propose an inductive fusion architecture which fuses planar LIDAR with RGB images to complete the depth map of the surrounding environment. %
The proposed inductive fusion block optimally combines the dense context features from RGB images with the sparse depth features from planar LIDAR. %
Our method outperforms conventional methods on the NYUDepthV2 dataset and KITTI odometry dataset. We also achieve promising results on the simulated planar LIDAR dataset. %
Further work will apply the dense depth image generated by the proposed architecture to perception tasks such as object detection and vehicle heading estimation. %
We will also apply the proposed method to real planar LIDAR depth completion tasks.%

\section{Acknowledgments}
This work was supported by the CMU Argo AI Center for Autonomous Vehicle Research.
\bibliographystyle{IEEEtran}

\bibliography{ref}

\end{document}